\documentclass[conference]{IEEEtran}
\usepackage{cite}
\usepackage{hyperref}
\usepackage{amsmath,amssymb,amsfonts}
\usepackage{algorithmic}
\usepackage{graphicx}
\usepackage{textcomp}
\usepackage{xcolor}
\def\BibTeX{{\rm B\kern-.05em{\sc i\kern-.025em b}\kern-.08em
    T\kern-.1667em\lower.7ex\hbox{E}\kern-.125emX}}
\begin{document}

\title{Data Augmentation for Emotion Detection in \\Small Imbalanced Text Data}
\author{\IEEEauthorblockN{Anna Koufakou,  Diego Grisales, Ragy Costa de jesus, Oscar Fox}
\IEEEauthorblockA{\textit{Department of Computing and Software Engineering, Florida Gulf Coast University, Fort Myers, Florida, USA} \\
\texttt{akoufakou@fgcu.edu}}
}

\maketitle

\begin{abstract}
Emotion recognition in text, the task of identifying emotions such as joy or anger, is a challenging problem in NLP with many applications. One of the challenges is the shortage of available datasets that have been annotated with emotions. Certain existing datasets are small, follow different emotion taxonomies and display imbalance in their emotion distribution. In this work, we studied the impact of data augmentation techniques precisely when applied to small imbalanced datasets, for which current state-of-the-art models (such as RoBERTa) under-perform. Specifically, we utilized four data augmentation methods (Easy Data Augmentation EDA, static and contextual Embedding-based, and ProtAugment) on three datasets that come from different sources and vary in size, emotion categories and distributions. Our experimental results show that using the augmented data when training the classifier model leads to significant improvements. Finally, we conducted two case studies: a) directly using the popular chat-GPT API to paraphrase text using different prompts, and b) using external data to augment the training set. Results show the promising potential of these methods.
\end{abstract}


\section{Introduction}
Today's society has been tremendously impacted by recent advances in machine learning and Natural Language Processing (NLP) in particular. There is great interest to apply current state-of-the-art models to a myriad of data and tasks. One of the research areas that has attracted interest is the task of detecting emotions expressed in text. There are many applications from intelligent recommender systems to social media monitoring to mental health intervention. For existing work in this area, the reader is referred to reviews such as 
\cite{nandwani2021review}. 

There are several challenges related to NLP for emotion recognition. Researchers must handle datasets in different formats (for example, a traditional essay or dialogue exchange or social media post). Emotions may be hard to comprehend by humans in the first place, so they are complicated for human annotators as well. Different taxonomies for categorizing emotions exist, e.g., Ekman \cite{ekman1992argument} or Plutchik \cite{plutchik1984emotions}. All this is further complicated by the fact that, the same paragraph or even social media post may express more than one emotion. This shows the inherent difficulty in collecting and annotating large datasets for emotion recognition. Several datasets that do exist contain a small number of records: for example, a few thousand in COVID-19 survey dataset from \cite{kleinberg2020measuring}. Furthermore, these datasets are heavily dominated by one or two emotions, making the datasets imbalanced and therefore harder for algorithms to detect the rare emotion categories. 

One avenue that has been pursued to help alleviate the scarcity of data generally in NLP is Data Augmentation. This term refers to artificially increasing the size of the training data by generating new records from the existing training ones. There are various techniques that have been proposed for NLP, from replacing random words in the text with their synonyms \cite{wei2019eda} to translating the record to another language and then back to the original language \cite{mallinson-etal-2017-paraphrasing} to using sophisticated models for paraphrasing the text in the original records \cite{dopierre-etal-2021-protaugment}.

Recently, there have been thorough surveys of data augmentation for NLP, e.g. \cite{feng2021survey, 
bayer2022survey}. In this paper, we followed \cite{pellicer2023data}, which conducted an extensive comparison of augmentation techniques for NLP. The authors showed that augmentation works for different datasets, including an emotion-based dataset. 
Additionally, researchers have applied data augmentation for mental health classification on social media data \cite{ansari2021data}.  They used Logistic regression, Support Vector Machine and Random Forest classifiers on two datasets labeled for stress and for depression/suicide. Augmentation for emotion recognition in data from software repositories was successfully used in \cite{imran-emo-software-2023}. On the other hand, negative findings have been reported regarding the effectiveness of data augmentation on downstream classification, see \cite{longpre2020effective} and \cite{okimura2022impact}. Very recently, three teams in the WASSA-23 shared tasks, based on a more recent version of one the datasets in our paper, used data augmentation \cite{barriere-etal-2023-findings}. Nevertheless, our work is the first to provide the community with a comprehensive comparative study of Text Data Augmentation, particularly tailored to emotion detection. 
Specifically, in this work, our goal is to explore the impact of data augmentation on the classification of emotions in small imbalanced data. We \textit{intentionally} selected recent emotion-labeled datasets that have been shown to have low accuracy in emotion detection tasks (for example, see \cite{koufakou2022automatically,plaza-del-arco-etal-2020-emoevent}), specifically because of the larger potential benefits from augmentation. Our datasets include a 
dataset based on a UK survey related to COVID-19 \cite{kleinberg2020measuring}; one collected from social media posts 
\cite{plaza-del-arco-etal-2020-emoevent}; and one 
containing essays written after reading news articles \cite{tafreshi2021wassa} 
 (we describe the datasets in detail in Section \ref{sec-data}). 

\begin{table*}
\centering
\caption{Dataset Comparison: Count of records per emotion category for each dataset, divided in the train and test sets we used, total records, and average length of records in the entire dataset. A ``--'' indicates the emotion is not present in the Dataset. }
\label{tab-data}
\centering
\begin{tabular}{lrrrrrrrr}
\hline
 & \multicolumn{2}{c}{~COVID-19 \cite{kleinberg2020measuring}~} & & \multicolumn{2}{c}{EmoEvent-EN \cite{plaza-del-arco-etal-2020-emoevent}} 
 & &  \multicolumn{2}{c}{~~WASSA-21 \cite{tafreshi2021wassa}~~} \\
\cline{2-3} \cline{5-6} \cline{8-9} 
\textit{\textbf{Emotion}}        & 
Train   & ~Test && 
Train   & ~Test &&
Train   & ~Test \\
\hline
\textbf{Anger}          & 75   & 32  && 274  & 118  
&& 349  & 122\\
\textbf{Anxiety}        & 966  & 415 && --    & --   
&& --    & --  \\
\textbf{Disgust}        & --   & --  && 536 & 229 
&& 149  & 28  \\
\textbf{Fear}           & 161  & 69  && 106  & 45  
&& 194  & 70 \\
\textbf{Joy}            & --    & --  && 1,427  & 612 
&& 82   & 33 \\
\textbf{Neutral/other } & --    & --   && 2,313  & 992 
&&  275  & 55 \\
\textbf{Relaxation}     & 233  & 100 && --   & --   
&& --    & --  \\
\textbf{Sadness}        & 250  & 107 && 291 & 125  
&& 647  & 177 \\
\textbf{Surprise}       & --    & --   && 165  & 70  
&& 164  & 40 \\
\hline
\textbf{\textit{Total records}} & 1,685 & ~~723 && 5,112 & 2,191 
&&1,860 & ~~525\\
\hline
\textbf{\textit{Avg. record length}} & \multicolumn{2}{c}{633} 
&& \multicolumn{2}{c}{138} 
&& \multicolumn{2}{c}{443}\\ 
\hline
\end{tabular}
\end{table*}

Our contribution is that we applied augmentation specifically focused to emotion datasets with a small size and imbalanced class distribution. 
For example, it has been shown in \cite{mundra-etal-2021-wassa, koufakou2022automatically} that using records from a larger emotion-annotated dataset, GoEmotions \cite{demszky2020goemotions}, alongside existing training data can improve classification performance for data we used in this paper; however, they did not perform any augmentation techniques on the original training records. Additionally, it was shown in \cite{koufakou2022automatically} that classification on their GoEmotions-based set \cite{demszky2020goemotions} had relaively high f1-macro's, therefore we did not try to augment and classify GoEmotions in our experimental study. Similarly for an 
emotion-labeled dataset used in \cite{pellicer2023data}.

The organization of this paper is as follows: Section \ref{sec-data} contains a description of the datasets used in this paper. Section \ref{sec-methods} gives a background on data augmentation methods for NLP, focusing on methods we used for this work. Then, it describes the overall process we used for augmenting our data and for classification. Section \ref{sec-results} presents our experimental setup, results and observations. Finally, Section \ref{sec-conclusion} includes concluding remarks and future research.

\section{Datasets}
\label{sec-data}
In this paper, we experimented with the datasets described in the following. We focused on datasets that were small and imbalanced that were also presented in 2020 or later, rather than earlier datasets such as SemEval-2018 Task 1 \cite{mohammad-etal-2018-semeval}. 
The datasets have a variety of emotions and  distributions, and come from different sources such as social media or essays written by survey participants. All datasets are in English. 

\textbf{COVID-19 Survey Data}\footnote{Data available at \url{https://github.com/ben-aaron188/\\covid19worry}} was presented in \cite{kleinberg2020measuring}. This data was collected via a survey in the UK under lockdown for COVID-19 (in 2020). The participants in that survey wrote a paragraph of text as well as entered demographic data (e.g. gender) and ratings for several emotions. Following \cite{koufakou2022automatically}, we selected to use the ``chosen emotion'': a category chosen by each participant out of several emotion options. 
As in \cite{koufakou2022automatically}, we kept the emotion categories with at least 4\% of the total records, which resulted in a dataset of 2,408 records. This dataset does not follow Ekman \cite{ekman1992argument} or other emotion taxonomy. The resulting dataset contains Anger, Anxiety (dominant), Fear, Relaxation, and Sadness. For our experiments, we chose to use stratified split (70-30) to divide the records into a train and a test set (we did this for all datasets to which this applied). 

\textbf{EmoEvent-EN Data}\footnote{Data available at \url{https://github.com/fmplaza/EmoEvent}} was presented in \cite{plaza-del-arco-etal-2020-emoevent}. The dataset was collected from the Twitter platform based on various events that took place in April 2019 (for example, the Notre Dame Cathedral Fire). In order to select affective tweets, the authors in \cite{plaza-del-arco-etal-2020-emoevent} used Linguistic Inquiry and Word Count (LIWC) \cite{pennebaker2001linguistic}. The selected tweets were then annotated by Amazon MTurkers using one of seven emotions: six emotions from Ekman's taxonomy \cite{ekman1992argument} plus ``neutral or other emotions''. In addition, each record was labeled as offensive or not, which we ignored for this paper. The total records in the resulting sets were 7,303 in English (there was also a dataset in Spanish, which we leave for future research). 
The dataset on the online repository had already replaced any hashtags, urls, or mentions with `HASHTAG', `URL' and `USER'. We split the data into train and test using a stratified split (70-30).

{\textbf{WASSA-21 Data}}\footnote{Data available at \url{https://competitions.codalab.org/competitions/28713}} was part of the 11th Workshop on Computational Approaches to Subjectivity, Sentiment \& Social Media Analysis (WASSA 2021) Shared Task on Empathy Detection and Emotion Classification summarized in \cite{tafreshi2021wassa}. This dataset is an extension of \cite{buechel-etal-2018-modeling}'s dataset based on news articles related to harm to an individual, group, nature, etc. The dataset contains essays written to express the author's empathy and distress in reaction to the news articles. The essays are annotated for empathy and distress, as well as personality traits and demographic information (age, gender, etc.). Each essay is also tagged with one of the Ekman’s emotions \cite{ekman1992argument}: Anger, Disgust, Fear, Joy, Sadness, and Surprise. We only focused on the emotion for each essay, not the empathy or distress labels, following \cite{koufakou2022automatically}. We did not perform train-test split for this data, as the WASSA-21 dataset already had distinct sets for training (1,860 records) and testing (525 records). 

A comparison is shown in Table \ref{tab-data}: this shows number of records per each emotion, if the emotion is present in that dataset. The counts are divided into train and test sets. The last two rows of the Table display the total number of records in each set followed by the average length of the records in characters. As shown in Table \ref{tab-data}, there is variety of emotion emotion categories as well as distributions. For example, COVID-19 is heavily skewed towards Anxiety (57\%), which is not part of the Ekman taxonomy \cite{ekman1992argument}, therefore not found in the rest of the datasets. WASSA-21 is dominated by Anger and Sadness, emotions found in all datasets. In contrast, EmoEvent-EN 
is heavily dominated by Joy and Others (around 73\% for these two emotions) so it is more positive.  
As far as size, two datasets (COVID-19 and WASSA-21) have around 2.5K records, while EmoEvent-EN is more than 3 times larger (around 7K records). On the other hand, the EmoEvent-EN dataset contains tweets, so the length of each record is shorter, while the other two datasets have essays (paragraphs of sentences) - see average length in characters in Table \ref{tab-data}. 

\section{Methodology}
\label{sec-methods}

\subsection{Background: Data Augmentation Methods}
In this section, we provide background for all the data augmentation methods we used in this work. We chose widely-used augmentation methods for NLP based on simple word manipulations or more sophisticated based on the entire record.
All these methods have been shown to perform well (e.g. see \cite{pellicer2023data}) therefore we chose to use them for our work.

\textbf{Easy Data Augmentation (EDA)}: This simple technique proposed in \cite{wei2019eda} has been shown to be quite successful. EDA combines several operations for the words in the original record. For any given sentence in the training records, EDA randomly chooses and performs one of the following (stop words are not considered, and there is a parameter for choosing the percent of the words to be altered):
\begin{itemize}
\item Synonym Replacement: random words are replaced
by random synonyms from a dictionary e.g. WordNet\cite{miller1995wordnet};
\item Random Insertion: a random word from the given sentence
is chosen, and one of its synonyms is then inserted in the sentence at a random position;
\item Random Swap: randomly chosen words are swapped;
\item Random Deletion: random words are removed.
\end{itemize}

\textbf{Embeddings}: Rather than using synonyms from a dictionary as in the previous technique, one could employ word \textit{embeddings} instead. 
These methods start by representing the words by embeddings, or vectors of $n$-dimensions. Then, they replace or insert words that are found to be \textit{similar} in the word embedding space, e.g. by using cosine distance. In earlier work, for example, \cite{wang2015s}, researchers used \textit{static} pre-trained word embeddings. In order to take advantage of the more recent advances, we can use \textit{contextual} embeddings, for example generated by a pre-trained transformer model such as BERT (Bidirectional Encoder Representations
from Transformers) \cite{devlin2019bert}. In summary, the word in the original text $w$ is replaced by words predicted by a model such as BERT based on the context around $w$ in the original text. These methods may also chose random words to replace, insert etc. as in the previous parapraph.

\textbf{BART Paraphraser ProtAugment}:
Instead of focusing on individual words, ProtAugment \cite{dopierre-etal-2021-protaugment} paraphrases the original text. This technique employs BART, a model ``combining Bidirectional and Auto-Regressive
Transformers" \cite{lewis-etal-2020-bart}, to generate paraphrases. The authors fine-tuned BART on the paraphrase generation task using various datasets. They also utilized Diverse Beam Search \cite{vijayakumar2018diverse} and Back Translation \cite{mallinson-etal-2017-paraphrasing} to help generate diverse outputs.

\subsection{Overall Process for Augmentation and Classification}
\label{sec-overall-process}
First, the dataset was split into training and testing sets if needed (Section \ref{sec-data} has the details for each set). The training data was fed as input to each of the Data Augmentation techniques (described in the previous paragraphs). Using each specific augmentation method, we generated a new record $x'$ for each original training record $x$. For this new record $x'$, we also made a label copy $y'$, where we copied the original training label $y$. In all our experiments, we generated five paraphrased records for each original record. 
At the end of this augmentation phase, we had a new `augmented' training set with text and labels. We then concatenated the `augmented' training set to the original training set and we presented this `increased' training dataset to the classifier during the training phase. This set of records then was transformed into features for training the classification model. The test set was also transformed into features and fed into the trained classifier in order to predict the test labels.

\section{Experimental Results}
\label{sec-results}

\subsection{Experimental Setup}
We run our code using Google colab\footnote{\url{https://colab.research.google.com}} (our code is at \url{https://github.com/A-Koufakou/AugEmotionDetection}). For augmentation, we followed code provided by \cite{pellicer2023data} online\footnote{\url{https://github.com/lucasfaop/survey\_text\_augmentation}}, which used an NLP augmentation library called NLPAug\footnote{\url{https://github.com/makcedward/nlpaug}}. 
For any choices, we used the same as \cite{pellicer2023data}: for example, for Embeddings, we used 200-dimensional Glove Embeddings \cite{pennington2014glove} or BERT for Contextual Embeddings. For our classification, we used Simple Transformers\footnote{\url{https://simpletransformers.ai}}, a library built on Hugging Face\footnote{\url{https://huggingface.co/transformers}}. 
Since RoBERTa (Robustly optimized BERT approach), a transformer-based model \cite{liu2019roberta}, was shown to outperform other deep learning models in \cite{koufakou2022automatically} for two of our datasets (COVID-19 and WASSA-21), we chose to use the RoBERTa model, specifically, \texttt{roberta-base}. 
For every experiment, we fine-tuned the model on our data, so that the pre-trained model can further learn from our data. We used 2 Epochs, learning rate of $1e^{-5}$, maxlen of 256 and batch of 8, based on early trials and \cite{koufakou2022automatically}. 
We repeated each experiment (train/test with each set) 3 times and then reported the average performance on the predicted test labels.
We reported results based on:

\begin{equation}
\small
Precision ~(or Recall) = \frac{TP}{TP+FP ~(or FN)}
\end{equation}
\begin{equation}
\small
Accuracy = \frac{TP+TN}{N}
\end{equation}
\begin{equation}
\small
f1\textnormal{-}score=\frac{2 \times Precision \times Recall}{Precision+Recall}
\end{equation} 
where $TP$ is True Positives, $FP$ False Positives, $FN$ False
Negatives, and $N$ total number of records. We presented macro-averaged Precision, Recall and f1: for example, f1-macro averages the f1-score over all classes, and is well-suited to imbalanced class distributions \cite{koufakou2022automatically,pellicer2023data}.

\subsection{Results}
Our classification results with the original and the augmented train datasets are shown in Table \ref{tab-data}. 
Each of the values in the Table is an average of 3 runs. The standard deviation of the results ranges from 0.01 to 0.04 depending on the dataset.

\begin{table}[t]
\centering
\caption{Results of training with original vs augmented train sets. Bold numbers indicate maximum. f1, P, R are macro-averaged.}
\label{tab-results}
\begin{tabular}{llcccc}
 \hline
\textbf{DataSet} & \textbf{Training} & \textbf{Acc} & \textbf{f1} & P & R \\ \hline
COVID-19  & Original  & \textbf{61.55} & 31.09 & 31.03 & 32.51 
\\
& EDA & 58.92 & \textbf{45.51} & \textbf{46.89} & \textbf{44.51} \\
& Embed & 59.11 & 44.80 & 45.86 & 44.13 
\\
& BERT Embed  & 59.20 & 44.44 & 46.50 & 43.04
\\
& ProtAug  & 58.23	& 44.08	& 45.34 &	43.31
\\
\hline
EmoEvent-EN & Original  & \textbf{62.45} & 37.77 \        & 42.40 & 38.13\\
& EDA      & 60.78 & 47.74         & 49.86 & 46.38 \\
& Embed    & 61.69 & 48.28          & 50.21 & 46.84 \\
& BERT Embed   & 62.04 & \textbf{49.65} & \textbf{53.58}& \textbf{47.28}\\
& ProtAug & 59.84	& 46.88	& 48.08	& 45.86\\
\hline
WASSA-21 & Original & 59.11 & 38.07 & 45.76 &  39.49\\
&EDA  & \textbf{62.92} &  \textbf{54.79} & \textbf{56.11} & \textbf{54.47} \\
&Embed & 61.65 & 53.50 & 54.95 & 53.05 \\
&BERT Embed & 62.54 & 54.44 & 56.07 & 54.27\\
&ProtAug &	60.63 &	53.09	& 54.74 &	52.95\\
\hline
\end{tabular}
\end{table}

There was no single winner augmentation technique; we observed that the EDA method did the best for data that contains essays written by authors given a specific question or topic (COVID-19 and WASSA-21), while the embeddings methods and especially contextual embeddings were the top performing methods for tweet-based data (EmoEvent). In more detail, in Table \ref{tab-results} for COVID-19, EDA had the best f1-macro (45.51\%), improving the original f1-macro by about 14\%. The rest of the methods followed around 44\%. In Table \ref{tab-results} for WASSA-21, we see that EDA is once again the winner (54.79\% f1-macro) improving on the non-augmented f1-macro by almost 17\%. The other methods followed closely behind.\footnote{Note: the winning team on the related WASSA-21 task \cite{mundra-etal-2021-wassa} achieved f1-macros in mid/upper 50's, with ensembles of several pre-trained transformers each fine-tuned on WASSA-21, and augmentation with GoEmotions \cite{demszky2020goemotions} (see later sections). Exploring ensembles is left for future research.} In contrast, Table \ref{tab-results} shows that the BERT Embeddings Augmentation did the best for EmoEvent-EN (49.65\%) improving on the non-augmented f1-macro by almost 12\%. The static (Glove) Embedding method followed closely, with the rest of the augmenters in 1 or 2\% lower. 

\begin{figure}[b] 	
\vspace{-6mm} 
\centering 	
\includegraphics[scale=0.41]{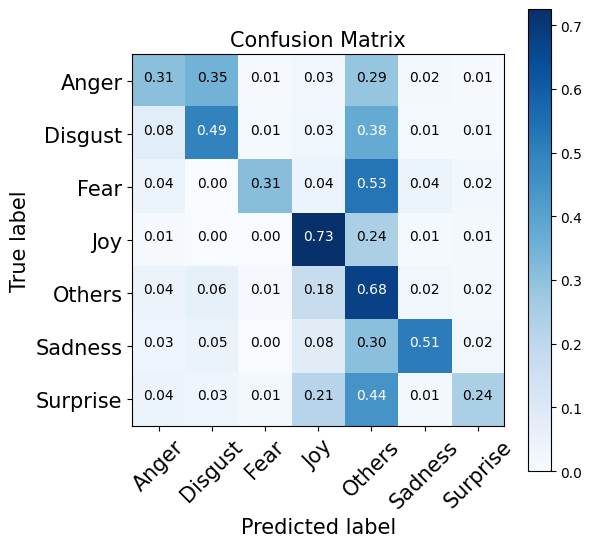}
\caption{Example Confusion Matrix (EmoEvent-EN, BERT Augmentation)} 	
\label{fig:fig-conf} 
\end{figure}

In Table \ref{tab-results}, augmented-based f1-macro's were always higher than non-augmented, however the same was not true for accuracy: for COVID-19 and EmoEvent-EN, accuracy was higher for non-augmented training than augmented. We examined the predictions per class, and when the model was trained with only original training, it did very well on the dominant class (Anxiety, for example in COVID-19) but much worse in the smaller classes (Anger and Fear, in COVID-19) - see Table \ref{tab-data} for emotion distributions. The EDA results showed better performance overall, but did mislabel more records for Anxiety, which made accuracy lower.
To examine the per-class results further, see the confusion matrix for EmoEvent-EN in Fig. \ref{fig:fig-conf}. We selected BERT Augmentation as it does the best for EmoEvent-EN (see Table \ref{tab-results}). From the confusion matrix, we can see that the model did relatively well on predicting Joy and Others (dominant emotions). The model confused Anger, Disgust, Sadness, and Surprise many times with Others. These issues were observed in the original paper \cite{plaza-del-arco-etal-2020-emoevent}: for example, the annotators had trouble annotating Fear, Disgust and Surprise, or to distinguish between Anger and Disgust (complementary emotions). 

In order to examine the strength of the data augmentation, we used the BLEU (Bilingual Evaluation Understudy) metric \cite{papineni-etal-2002-bleu} as per previous work \cite{ansari2021data,pellicer2023data}. BLEU is used for translation problems: lower BLEU score means more diversity from the original text, therefore a stronger augmentation. The BLEU scores for the augmented datasets in this work are shown in Table \ref{tab-blue}. We calculated BLEU score \textit{without brevity punishment} to avoid giving shorter texts an advantage per \cite{pellicer2023data}. As shown in Table \ref{tab-blue}, the top methods (EDA and BERT) had the lowest BLEU for EmoEvent and WASSA-21 respectively, meaning high lexical diversity could have helped these methods win. The situation is not the same for COVID-19: ProtAugmenter, not EDA, had the lowest score. 

\begin{table}[t]
\centering
\caption{Comparison of augmentation methods using BLEU score. Lower BLEU means higher lexical diversity, bold denotes minimum.}
\label{tab-blue}
\begin{tabular}{lrrr}
\hline
 & \textbf{COVID-19} & \textbf{EmoEvent-EN} & \textbf{WASSA-21} \\
\hline
EDA Aug          & 66.30           & 31.92         & \textbf{56.03}    \\
Embed Aug        & 75.41           & 40.35         & 67.21    \\
BERT Embed Aug   & 64.56           & \textbf{21.83}& 56.82    \\
Prot Aug         & \textbf{57.69}  & 76.20         & 72.73   \\
\hline
\end{tabular}
\end{table}

\begin{table*}[h]
\centering
\caption{Example of original records from COVID-19 and their paraphrasings}
\label{tab-example}
\begin{tabular}{ll} 
\hline
\textbf{Augmenter} & \textbf{Text}\\
\hline
\textbf{Original} & I am fairly safe in a secluded location, however I am very worried about the safety of my family and friends that live in big cities \\
\textbf{EDA} &	I am fairly safe in a sequestered location, however I be  rattling worried about the safety of my family and champion that live in  big city\\
\textbf{chatGPT} & I am grateful to be in a safe location, but my heart goes out to my loved ones who are suffering in cities\\
\hline
\textbf{Original} & I am unsure of what future holds worried I won’t see family members again and do things I enjoyed doing prior to virus\\
\textbf{EDA} & I am unsure of what holds future disquieted won’t see family members over again and do things I doing prior to virus \\
\textbf{chatGPT} & I am worried about what the future holds. I am worried I will not be able to see my family members again \\
& or do the things I  enjoyed doing prior to the virus\\
\hline
\end{tabular}
\end{table*}

Finally, regarding runtime and resource expectations of the augmentation techniques, we observed that EDA had low runtime and no GPU requirement in contrast to the other methods. 
Depending on the data,
we observed 1-3 minutes of runtime for EDA (no GPU) versus 28-35 minutes for ProtAugmenter and 2-3 hours for Embedding Augmentation methods (static embeddings were slightly faster than contextual), using GPU.

\subsection{Case Study A: Paraphrasing with chatGPT}

Motivated by the improvements achieved by the data augmentation using existing techniques already explored in the literature (e.g. \cite{pellicer2023data,ansari2021data}), we were interested to directly experiment with the popular chatGPT (Chat Generative Pre-Trained Transformer). This is a Large Language Model-based chatbot developed by OpenAI, which made headlines having reached 1 million users in a few days. We were only able to conduct a case study with COVID-19, due to lack of time and space. We also saw work just presented in WASSA-23 \cite{lu-etal-2023-hit}: the HIT-SCIR team competed in WASSA-23 shared tasks and also used chatGPT paraphrasing for augmenting the WASSA-23 train sets (using RoBERTa and bi-LSTM for their emotion prediction model). Further exploring their research as well as expanding our chatGPT work to more datasets is a focus of our future research. 

We wrote a script to connect to the OpenAI chat API\footnote{\url{https://platform.openai.com/docs/api-reference/chat}} and explored various prompts for paraphrasing training records. We used a single prompt-response cycle for each record. The prompts with which we experimented that showed promise were `summarize the following in the first person' (denoted as chatGPT-summ) and `using a sympathetic tone, paraphrase the following' (denoted as chatGPT-symp). 
In summary, paraphrasing using a `summarize' prompt in chatGPT (chatGPT-summ) had the highest f1-macro (46.10\%) followed closely by EDA (45.51\%). The `sympathize' prompt in chatGPT (chatGPT-symp) had a lower f1-macro (44.72\%). We tried another model we saw in the literature: a T5 model trained on ChatGPT paraphrase data\footnote{\url{https://huggingface.co/humarin/chatgpt_paraphraser_on_T5_base}}, with lower f1-macro results.

Finally, we looked at example paraphrases displayed in Table \ref{tab-example}. We only show results for EDA and chatGPT due to space. In Table \ref{tab-example}, the paraphrases from chatGPT have a better structure and vary from the original in words and phrasing. For example, it  paraphrased `fairly safe in a secluded location' to `grateful to be in a safe location'. This is also supported by the BLEU score for chatGPT paraphrases: it is much lower (25.04) than the values shown for COVID-19 in Table \ref{tab-blue}. This shows that chatGPT paraphrased data were more lexically diverse than the original data compared to the other augmenters. EDA results on the other hand inserted random words, such as `rattling', replaced `friends' with `champion', and swapped the order of the words in the `future holds'. Nevertheless, EDA is a simple method non-requiring GPU or API calls with a very low runtime comparably. 

\subsection{Case Study B: Augmentation using external data}

Besides augmentation techniques that reword or paraphrase the original data, recent work has proposed augmenting with an external dataset, perhaps from a different domain or source. For example, both \cite{mundra-etal-2021-wassa} and \cite{koufakou2022automatically} used GoEmotions \cite{demszky2020goemotions} (human annotated reddit comments): the first for augmenting WASSA-21, and the second for both COVID-19 and WASSA-21. As both papers showed improvements, we wanted to compare our results against augmenting these data with GoEmotions. First, to match our data format, we selected a subset of the GoEmotions train set based on records annotated with a single Ekman label (about 4,300 records). For each experiment, we only kept the emotions that were present in the original dataset. Finally, similar to Section \ref{sec-overall-process}, we concatenated that data with the original train set, and proceeded as before. 
The results were favorable for WASSA-21: the model trained with the GoEmotions-augmented data resulted in 55.94\% f1-macro (surpassing EDA's 54.79\% in Table \ref{tab-results}). However, for the COVID-19 data, f1-macro was 35.74\% (much lower than 45.51\% for EDA in Table \ref{tab-results}). Our interpretation is that WASSA-21's emotions are more representative of the Ekman taxonomy \cite{ekman1992argument}, therefore better suited to the additional records from GoEmotions. On the other hand, COVID-19 is heavily influenced by Anxiety, and the extra records we added from GoEmotions for Anger, Fear and Sadness led to negligible improvement in the prediction of these emotions. Nevertheless, we feel this topic warrants more in-depth exploration in our future work.

\section{Conclusions}
Detecting emotions such as joy or sadness in text is a significant area of research with many challenges. An important challenge is the availability of large annotated data. In this work, we experimented with several data augmentation techniques that have been successfully utilized in the literature, to explore potential improvements in emotion detection. We experimented with various emotion-labeled datasets, originating from different sources (e.g. social media posts versus essays) and varying in topics, emotions, distributions etc. Our experiments showed that using the augmented datasets for training led to large improvements: the best performing method in each dataset improved the f1-macro by 12-16\%. A simple technique, EDA, performed the best on essay data, such as WASSA-21, while contextualized embeddings did the best for tweet-based data. Additionally, we performed two case studies: a) using chatGPT, a very popular pre-trained large language model, to paraphrase the training set with different prompts, and b) using external data (GoEmotions) to augment the training set. 
The experiments show the promising potential of these techniques. 
Finally, we examined the strength of the resulting augmented text (lexical diversity vs original, using BLEU score) as well as the execution runtime. Paraphrasing with chatGPT, for example, was much more lexically diverse than simpler techniques (EDA), at the cost of much higher runtime and complexity.  
For future work, we intend to explore the impact of changing certain (hyper)parameters as well as experiment with more data (e.g. the latest WASSA sets presented in 2023 \cite{barriere-etal-2023-findings}), augmentation methods and ensembles. 
We also aim to methodically study `prompt engineering' in chatGPT, as it has been shown \cite{liu-prompt-survey2023} that a well-designed prompt may vastly improve its results.

\label{sec-conclusion}

\bibliographystyle{./IEEEtran}

\bibliography{augment}

\begin{thebibliography}{10}
\providecommand{\url}[1]{#1}
\csname url@samestyle\endcsname
\providecommand{\newblock}{\relax}
\providecommand{\bibinfo}[2]{#2}
\providecommand{\BIBentrySTDinterwordspacing}{\spaceskip=0pt\relax}
\providecommand{\BIBentryALTinterwordstretchfactor}{4}
\providecommand{\BIBentryALTinterwordspacing}{\spaceskip=\fontdimen2\font plus
\BIBentryALTinterwordstretchfactor\fontdimen3\font minus \fontdimen4\font\relax}
\providecommand{\BIBforeignlanguage}[2]{{%
\expandafter\ifx\csname l@#1\endcsname\relax
\typeout{** WARNING: IEEEtran.bst: No hyphenation pattern has been}%
\typeout{** loaded for the language `#1'. Using the pattern for}%
\typeout{** the default language instead.}%
\else
\language=\csname l@#1\endcsname
\fi
#2}}
\providecommand{\BIBdecl}{\relax}
\BIBdecl

\bibitem{nandwani2021review}
P.~Nandwani and R.~Verma, ``A review on sentiment analysis and emotion detection from text,'' \emph{Social Network Analysis and Mining}, vol.~11, no.~1, pp. 1--19, 2021.

\bibitem{ekman1992argument}
P.~Ekman, ``An argument for basic emotions,'' \emph{Cognition \& emotion}, vol.~6, no. 3-4, pp. 169--200, 1992.

\bibitem{plutchik1984emotions}
R.~Plutchik, ``Emotions: A general psychoevolutionary theory,'' \emph{Approaches to emotion}, vol. 1984, no. 197-219, pp. 2--4, 1984.

\bibitem{kleinberg2020measuring}
B.~Kleinberg, I.~van~der Vegt, and M.~Mozes, ``Measuring emotions in the covid-19 real world worry dataset,'' in \emph{Proceedings of the 1st Workshop on NLP for COVID-19 at ACL 2020}, 2020.

\bibitem{wei2019eda}
J.~Wei and K.~Zou, ``Eda: Easy data augmentation techniques for boosting performance on text classification tasks,'' in \emph{Proceedings of the 2019 Conference on Empirical Methods in Natural Language Processing (EMNLP-IJCNLP)}, 2019.

\bibitem{mallinson-etal-2017-paraphrasing}
J.~Mallinson, R.~Sennrich, and M.~Lapata, ``Paraphrasing revisited with neural machine translation,'' in \emph{Proceedings of the 15th Conference of the {E}uropean Chapter of the Association for Computational Linguistics}, Valencia, Spain, Apr. 2017, pp. 881--893.

\bibitem{dopierre-etal-2021-protaugment}
T.~Dopierre, C.~Gravier, and W.~Logerais, ``{PROTAUGMENT}: Unsupervised diverse short-texts paraphrasing for intent detection meta-learning,'' in \emph{Proceedings of the 59th Annual Meeting of the Association for Computational Linguistics}, Aug. 2021, pp. 2454--2466.

\bibitem{feng2021survey}
S.~Y. Feng, V.~Gangal, J.~Wei, S.~Chandar, S.~Vosoughi, T.~Mitamura, and E.~Hovy, ``A survey of data augmentation approaches for nlp,'' in \emph{Findings of the Association for Computational Linguistics}, 2021.

\bibitem{bayer2022survey}
M.~Bayer, M.~Kaufhold, and C.~Reuter, ``A survey on data augmentation for text classification,'' \emph{ACM Comput. Surv.}, vol.~55, no.~7, pp. 1--39, 2022.

\bibitem{pellicer2023data}
L.~Pellicer, T.~Ferreira, and A.~Costa, ``Data augmentation techniques in natural language processing,'' \emph{Applied Soft Computing}, vol. 132, 2023.

\bibitem{ansari2021data}
G.~Ansari, M.~Garg, and C.~Saxena, ``Data augmentation for mental health classification on social media,'' in \emph{Proceedings of the 18th International Conference on Natural Language Processing (ICON)}, 2021.

\bibitem{imran-emo-software-2023}
M.~M. Imran, Y.~Jain, P.~Chatterjee, and K.~Damevski, ``Data augmentation for improving emotion recognition in software engineering communication,'' in \emph{Proceedings of the 37th IEEE/ACM International Conference on Automated Software Engineering}, 2023.

\bibitem{longpre2020effective}
S.~Longpre, Y.~Wang, and C.~DuBois, ``How effective is task-agnostic data augmentation for pretrained transformers?'' in \emph{Findings of the Association for Computational Linguistics: EMNLP 2020}, 2020.

\bibitem{okimura2022impact}
I.~Okimura, M.~Reid, M.~Kawano, and Y.~Matsuo, ``On the impact of data augmentation on downstream performance in natural language processing,'' in \emph{Proceedings of the 3rd Workshop on Insights from Negative Results in NLP}, 2022, pp. 88--93.

\bibitem{barriere-etal-2023-findings}
V.~Barriere, J.~Sedoc, S.~Tafreshi, and S.~Giorgi, ``Findings of {WASSA} 2023 shared task on empathy, emotion and personality detection in conversation and reactions to news articles,'' in \emph{Proceedings of the 13th Workshop on Computational Approaches to Subjectivity, Sentiment, {\&} Social Media Analysis}.\hskip 1em plus 0.5em minus 0.4em\relax Toronto, Canada: ACL, Jul. 2023, pp. 511--525.

\bibitem{koufakou2022automatically}
A.~Koufakou, J.~Garciga, A.~Paul, J.~Morelli, and C.~Frank, ``Automatically classifying emotions based on text: A comparative exploration of different datasets,'' in \emph{34th International Conference on Tools with Artificial Intelligence (ICTAI)}.\hskip 1em plus 0.5em minus 0.4em\relax IEEE, 2022.

\bibitem{plaza-del-arco-etal-2020-emoevent}
F.~{Plaza-del-Arco}, C.~Strapparava, L.~A. {Urena-Lopez}, and M.~T. {Martin-Valdivia}, ``\BIBforeignlanguage{English}{{{E}mo{E}vent: A Multilingual Emotion Corpus based on different Events}},'' in \emph{\BIBforeignlanguage{English}{Proceedings of the 12th Language Resources and Evaluation Conference}}, May 2020.

\bibitem{tafreshi2021wassa}
S.~Tafreshi, O.~De~Clercq, V.~Barriere, S.~Buechel, J.~Sedoc, and A.~Balahur, ``Wassa 2021 shared task: Predicting empathy and emotion in reaction to news stories,'' in \emph{Proceedings of the 11th Workshop on Computational Approaches to Subjectivity, Sentiment and Social Media Analysis}.\hskip 1em plus 0.5em minus 0.4em\relax Online: ACL, 2021.

\bibitem{mundra-etal-2021-wassa}
J.~Mundra, R.~Gupta, and S.~Mukherjee, ``{WASSA}@{IITK} at {WASSA} 2021: Multi-task learning and transformer finetuning for emotion classification and empathy prediction,'' in \emph{Proceedings of the 11th Workshop on Computational Approaches to Subjectivity, Sentiment and Social Media Analysis}.\hskip 1em plus 0.5em minus 0.4em\relax Online: ACL, Apr. 2021, pp. 112--116.

\bibitem{demszky2020goemotions}
D.~Demszky, D.~Movshovitz-Attias, J.~Ko, A.~Cowen, G.~Nemade, and S.~Ravi, ``Goemotions: A dataset of fine-grained emotions,'' in \emph{Proceedings of the 58th Annual Meeting of the Association for Computational Linguistics}, 2020, pp. 4040--4054.

\bibitem{mohammad-etal-2018-semeval}
S.~Mohammad, F.~Bravo-Marquez, M.~Salameh, and S.~Kiritchenko, ``{S}em{E}val-2018 task 1: Affect in tweets,'' in \emph{Proceedings of The 12th International Workshop on Semantic Evaluation}, Jun. 2018.

\bibitem{pennebaker2001linguistic}
J.~W. Pennebaker, M.~E. Francis, and R.~J. Booth, ``Linguistic inquiry and word count: Liwc 2001,'' \emph{Mahway: Lawrence Erlbaum Associates}, vol.~71, 2001.

\bibitem{buechel-etal-2018-modeling}
S.~Buechel, A.~Buffone, B.~Slaff, L.~Ungar, and J.~Sedoc, ``Modeling empathy and distress in reaction to news stories,'' in \emph{Proceedings of the 2018 Conference on Empirical Methods in Natural Language Processing}, Brussels, Belgium, Oct.-Nov. 2018, pp. 4758--4765.

\bibitem{miller1995wordnet}
G.~A. Miller, ``Wordnet: a lexical database for english,'' \emph{Communications of the ACM}, vol.~38, no.~11, pp. 39--41, 1995.

\bibitem{wang2015s}
W.~Y. Wang and D.~Yang, ``That’s so annoying!!!: A lexical and frame-semantic embedding based data augmentation approach to automatic categorization of annoying behaviors using\# petpeeve tweets,'' in \emph{Proceedings of the 2015 conference on empirical methods in natural language processing}, 2015, pp. 2557--2563.

\bibitem{devlin2019bert}
J.~Devlin, M.-W. Chang, K.~Lee, and K.~Toutanova, ``{BERT}: Pre-training of deep bidirectional transformers for language understanding,'' in \emph{Conference of the NACL: Human Language Technologies}, Jun. 2019.

\bibitem{lewis-etal-2020-bart}
M.~Lewis, Y.~Liu, N.~Goyal, M.~Ghazvininejad, A.~Mohamed, O.~Levy, V.~Stoyanov, and L.~Zettlemoyer, ``{BART}: Denoising sequence-to-sequence pre-training for natural language generation, translation, and comprehension,'' in \emph{Proceedings of the 58th Annual Meeting of the Association for Computational Linguistics}, Jul. 2020, pp. 7871--7880.

\bibitem{vijayakumar2018diverse}
A.~Vijayakumar, M.~Cogswell, R.~Selvaraju, Q.~Sun, S.~Lee, D.~Crandall, and D.~Batra, ``Diverse beam search for improved description of complex scenes,'' in \emph{Proceedings of the AAAI Conference on Artificial Intelligence}, vol.~32, no.~1, 2018.

\bibitem{pennington2014glove}
J.~Pennington, R.~Socher, and C.~D. Manning, ``Glove: Global vectors for word representation,'' in \emph{Proceedings of the 2014 conference on empirical methods in natural language processing (EMNLP)}, 2014.

\bibitem{liu2019roberta}
Y.~Liu, M.~Ott, N.~Goyal, J.~Du, M.~Joshi, D.~Chen, O.~Levy, M.~Lewis, L.~Zettlemoyer, and V.~Stoyanov, ``Roberta: A robustly optimized bert pretraining approach,'' \emph{arXiv preprint arXiv:1907.11692}, 2019.

\bibitem{papineni-etal-2002-bleu}
K.~Papineni, S.~Roukos, T.~Ward, and W.-J. Zhu, ``{B}leu: a method for automatic evaluation of machine translation,'' in \emph{Proceedings of the 40th Annual Meeting of the Association for Computational Linguistics}, 2002.

\bibitem{lu-etal-2023-hit}
X.~Lu, Z.~Li, Y.~Tong, Y.~Zhao, and B.~Qin, ``{HIT}-{SCIR} at {WASSA} 2023: Empathy and emotion analysis at the utterance-level and the essay-level,'' in \emph{Proceedings of the 13th Workshop on Computational Approaches to Subjectivity, Sentiment, {\&} Social Media Analysis}.\hskip 1em plus 0.5em minus 0.4em\relax Toronto, Canada: ACL, Jul. 2023, pp. 574--580.

\bibitem{liu-prompt-survey2023}
P.~Liu, W.~Yuan, J.~Fu, Z.~Jiang, H.~Hayashi, and G.~Neubig, ``Pre-train, prompt, and predict: A systematic survey of prompting methods in natural language processing,'' \emph{ACM Comput. Surv.}, vol.~55, no.~9, 2023.

\end{thebibliography}

\end{document}